\documentclass[11pt]{article} 

\usepackage[utf8]{inputenc} 

\usepackage{geometry} 
\usepackage{graphicx} 

\usepackage[singlelinecheck=false]{caption}

\usepackage{booktabs} 
\usepackage{array} 
\usepackage{paralist} 
\usepackage{verbatim} 
\usepackage{subfig} 
\usepackage{amsthm,amsmath}
\usepackage{amssymb}
\usepackage{float}
\usepackage{setspace}
\usepackage{xcolor}
\usepackage{hyperref}
\usepackage{dsfont} 

\usepackage{fancyhdr} 
\usepackage{sectsty}
\allsectionsfont{\sffamily\mdseries\upshape} 

\usepackage[nottoc,notlof,notlot]{tocbibind} 
\usepackage[titles,subfigure]{tocloft} 

\usepackage{array}
\usepackage{multirow}
\newcolumntype{L}[1]{>{\raggedright\let\newline\\\arraybackslash\hspace{0pt}}m{#1}}
\newcolumntype{C}[1]{>{\centering\let\newline\\\arraybackslash\hspace{0pt}}m{#1}}
\newcolumntype{R}[1]{>{\raggedleft\let\newline\\\arraybackslash\hspace{0pt}}m{#1}}

\definecolor{yellow}{RGB}{208, 218, 56}
\definecolor{green}{RGB}{120, 191, 61}
\definecolor{orange}{rgb}{0.99,0.69,0.07}

\usepackage{pifont} 
\usepackage{authblk}

\title{Random survival forests with multivariate longitudinal endogenous covariates}
\author[1]{Anthony Devaux}
\author[1]{Catherine Helmer}
\author[1,2,*,$\dagger$]{Robin Genuer}
\author[1,$\dagger$]{Cécile Proust-Lima}

\affil[1]{Univ. Bordeaux, INSERM, Bordeaux Population Health, UMR1219, Bordeaux, France}
\affil[2]{INRIA Bordeaux Sud-Ouest, Talence, France}
\affil[*]{Email: robin.genuer@u-bordeaux.fr}
\affil[$\dagger$]{These authors contributed equally to this work}
\date{January 2023} 

\begin{document}

\doublespacing

\maketitle

\textbf{Abstract:} Predicting the individual risk of a clinical event using the complete patient history is still a major challenge for personalized medicine. Among the methods developed to compute individual dynamic predictions, the joint models have the assets of using all the available information while accounting for dropout. However, they are restricted to a very small number of longitudinal predictors. Our objective was to propose an innovative alternative solution to predict an event probability using a possibly large number of longitudinal predictors. We developed \texttt{DynForest}, an extension of competing-risk random survival forests that handles endogenous longitudinal predictors. At each node of the tree, the time-dependent predictors are translated into time-fixed features (using mixed models) to be used as candidates for splitting the subjects into two subgroups. The individual event probability is estimated in each tree by the Aalen-Johansen estimator of the leaf in which the subject is classified according to his/her history of predictors. The final individual prediction is given by the average of the tree-specific individual event probabilities. We carried out a simulation study to demonstrate the performances of \texttt{DynForest} both in a small dimensional context (in comparison with joint models) and in a large dimensional context (in comparison with a regression calibration method that ignores informative dropout). We also applied \texttt{DynForest} to (i) predict the individual probability of dementia in the elderly according to repeated measures of cognitive, functional, vascular and neuro-degeneration markers, and (ii) quantify the importance of each type of markers for the prediction of dementia. Implemented in the R package \texttt{DynForest}, our methodology provides a novel and appropriate solution for the prediction of events from any number of longitudinal endogenous predictors. \\

\textbf{Keywords:} Individual dynamic prediction, Multivariate predictors, Random survival forest, Longitudinal data, Survival data, Competing risks.

\section{Background}

Quantifying the patient specific risk of disease or health events related to a disease progression based on patient's information has become a crucial issue in modern medicine. This may be done in order to monitor the disease progression, and adapt therapeutic strategies and medical choices according to their risk. One strategy is to predict the risk of event using only the data collected at the prediction time. However, in many contexts, patients data include repeated measures of markers which trajectories are highly predictive of the event. This is the case for instance with prostate specific antigen for the risk of prostate cancer recurrence \cite{proust-lima_development_2009} or serum creatinine for the risk of kidney graft failure \cite{fournier_dynamic_2019}. In other contexts, such as in cardiovascular disease, not only one specific marker but many potential markers may be relevant \cite{paige_landmark_2018}.

Longitudinal markers are endogenous variables in the sense that they may be affected by the event of interest \cite{rizopoulos_joint_2012}, and are usually measured intermittently with a measurement error. This makes their statistical analysis challenging. Three approaches were proposed in the literature for the prediction of a clinical event given longitudinal endogenous information: \textit{landmark approach} \cite{van_houwelingen_dynamic_2007}, \textit{joint models} \cite{proust-lima_development_2009} and \textit{regression calibration} techniques \cite{ye_semiparametric_2008}.

Landmark approach consists in considering only the subjects still at risk of the event at a prediction time $t$ (called landmark time) and including their information collected until $t$ to build a prediction tool for subsequent risk of event \cite{van_houwelingen_dynamic_2007, ferrer_individual_2019}. The longitudinal information of intermittently measured and prone-to-error markers up to $t$ can be included after a pre-processing step, by mixed models for instance \cite{proust-lima_development_2009, paige_landmark_2018, devaux_individual_2022, tanner_dynamic_2021}. In multivariate settings, the Cox model may be replaced by more advanced techniques, coming from statistical learning and adapted to survival data \cite{devaux_individual_2022, tanner_dynamic_2021} to account for the possibly large dimension of the predictors and their correlation. The landmark approach is relatively easy to implement and was shown to be robust to the misspecification of the marker trajectory or to the proportional assumption in the Cox model \cite{ferrer_individual_2019}. This makes it an appealing approach for extending the concept of individual dynamic prediction from a unique longitudinal marker to predictions from multivariate longitudinal markers. However, because it only relies on subjects at risk at the landmark time and exploits only the longitudinal information up to the landmark time, it suffers from a lack of efficiency, and is restricted to pre-determined prediction times.

When longitudinal and survival processes are inter-related as assumed in the dynamic prediction context, the joint modelling framework constitutes the most appropriate approach to handle this mutual dependence \cite{rizopoulos_joint_2012}. Joint models (JM) simultaneously model the longitudinal and survival processes over time while accounting for their association using shared latent quantities; and the posterior conditional individual probability of event given the longitudinal predictor history can be easily deduced. Initially developed for a single longitudinal predictor \cite{proust-lima_development_2009}, the method was then extended to a few longitudinal predictors \cite{hickey_joinerml_2018,rizopoulos_r_2016}. In contrast with landmark approaches, JM exploit all the available longitudinal information to build the prediction tool, thus leading to a better efficiency. However, their performances are very sensitive to the correct specification of the model \cite{ferrer_individual_2019}. Moreover, due to the complexity of their estimation, they are currently limited to a small number of longitudinal markers (usually 2 or 3) and thus cannot be used to predict individual risk of event in more complex settings \cite{hickey_joint_2016}.

In the context of a large to high number of longitudinal predictors, regression calibration (RC) techniques were proposed as an alternative to JM. RC is a two-stage approach which first summarizes the longitudinal predictors into time-fixed features as in the landmark approach but using all the repeated measures until the time of event or censoring, and then includes the features into prediction models. Several RC methods have been proposed with a first step using mixed models or functional data analysis \cite{yao_functional_2005} to summarize the multiple longitudinal predictors. Then, Cox model \cite{li_dynamic_2019}, penalized regression \cite{signorelli_penalized_2021} or random survival forests \cite{jiang_functional_2021,lin_functional_2021} have been used to derive the risk prediction. As in JM, RC techniques include all the available information on the markers and survival during the follow-up to build the prediction tool. However, they neglect the informative truncation of the longitudinal data due to the event, which can bias the estimates and impact the prediction accuracy \cite{albert_estimating_2010}.

In this work, we propose a novel methodology based on competing-risk random survival forests (RSF) \cite{ishwaran_random_2008, ishwaran_random_2014} to accurately predict a risk of event from possibly large-dimensional longitudinal predictors. RSF have become popular for prediction tasks as they can handle a high number of covariates and capture potentially complex associations. However, RSF have been limited so far to time-independent predictors. To extend RSF to intermittently measured and error-prone longitudinal predictors, we model them using linear mixed model. However, in contrast to the landmark and RC techniques, we directly incorporate those computations at each recursive step of the RSF tree building to better handle the informative truncation of the longitudinal endogenous data and thus provide more appropriate and accurate individual predictions.

The rest of this article is organized as follows. Section 2 introduces our extended random survival forest methodology, called \texttt{DynForest}, and describes how it can handle time-dependent endogenous predictors to predict a risk of event, possibly with multiple causes. Section 3 describes an extensive simulation study which aimed at validating \texttt{DynForest} methodology and at contrasting its performances with those of alternative approaches.
In section 4, \texttt{DynForest} is applied in a large population-based French cohort to predict the probability of experiencing a dementia before death from multiple markers stemming from neuropsychological evaluation, clinical evaluation and brain Magnetic Resonance Imaging (MRI). Finally, section 5 closes the work with a discussion.

\section{Methods}

\subsection{Framework and notations}

We consider a sample of $N$ subjects. For each individual $i \in \{1,...,N\}$, we denote $T_i$ the event time, $C_i$ the independent censoring time, and $\tilde{T}_i = \min{(T_i, C_i)}$ the observed time of event. We define $\delta_i$ the indicator of the cause of event with $\delta_i = k$ if subject experiences the event of cause $k \in \{1,...,K\}$ before censoring and $\delta_i = 0$ otherwise. We observe an ensemble $\mathcal{M}_x$ of $P$ time-independent covariates $X_{ip}$ ($p=1,...,P$), and an ensemble $\mathcal{M}_y$ of $Q$ time-dependent covariates $Y_{ijm}$ for $m=1,...Q$ measured at subject-and-covariate-specific times $t_{ijm}$ with $j=1...n_{im}$ the occasion and $t_{ijm} \leq \tilde{T}_i$.

Our methodology consists of a random survival forest (RSF) that incorporates an internal processing for handling time-dependent covariates. A RSF is an ensemble of $B$ survival decision trees that are ultimately aggregated together. Each tree $b \in \{1,...,B\}$ is built from a bootstrap sample of the original sample of $N$ subjects. This results, on average, in the exclusion of 37\% of the subjects that constitute the out-of-bag (OOB) sample, noted $OOB^b$.

\subsection{The tree building} 

\begin{figure}[H]
  \centering\includegraphics[width=\textwidth]{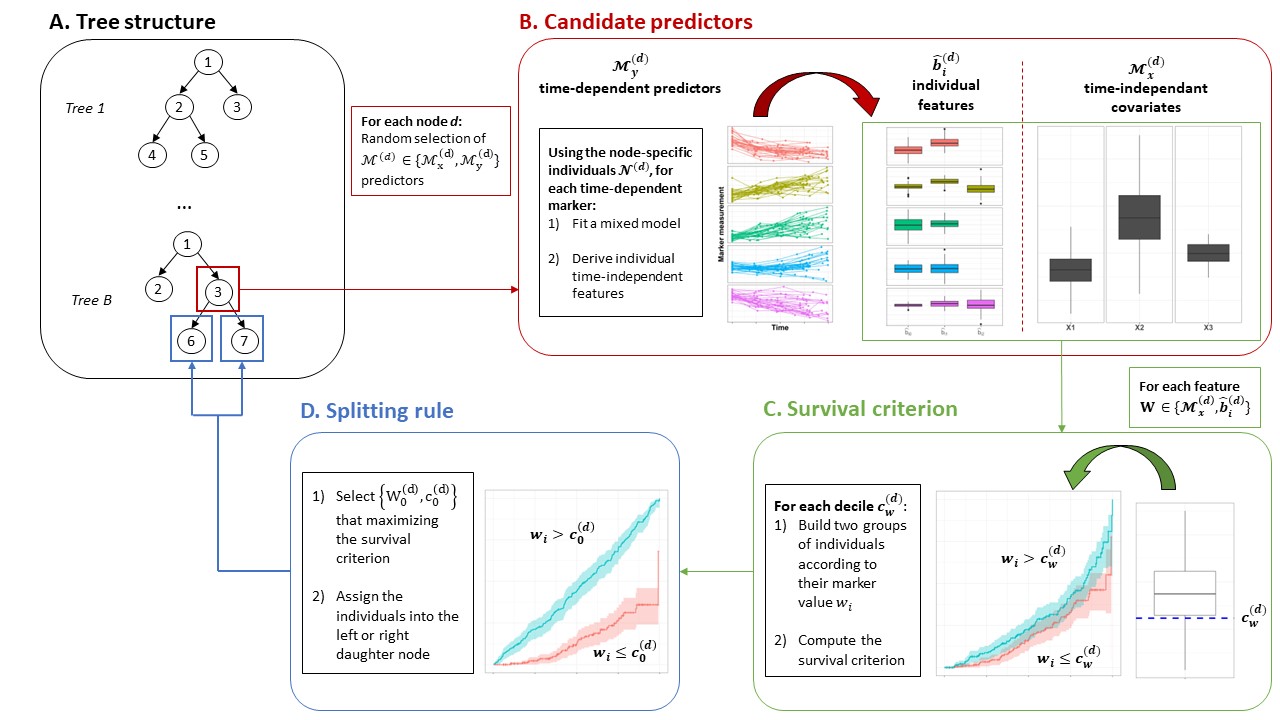}
  \caption{Overall scheme of the tree building in \texttt{DynForest} with (A) the tree structure, (B) the node-specific treatment of time-dependent predictors to obtain time-fixed features, (C) the dichotomization of the time-fixed features, (D) the splitting rule.}
  \label{fig:dynforest}
\end{figure}

A tree is a recursive procedure designed to partition the subjects into homogeneous groups regarding the outcome of interest. The overall tree building procedure is summarized in figure \ref{fig:dynforest}. Each tree recursively splits the bootstrap sample into two subgroups at junctions called nodes until the subgroups reach a minimal size. At each node $d \in \mathcal{D}$, the split is determined according to a dichotomized feature that maximizes the distance between the two groups; the distance definition depends on the nature of the outcome (see the Splitting rule section for the survival and competing risk settings). To improve accuracy and minimize the correlation between the trees, randomness is incorporated at each node $d$ by considering only a random subset of candidate covariates $\mathcal{M}^{(d)}=\{\mathcal{M}_x^{(d)},\mathcal{M}_y^{(d)}\} \subset \{\mathcal{M}_x,\mathcal{M}_y\}$ which size is a tuning parameter, called \textit{mtry}.

\subsubsection{Internal processing for time-dependent covariates}

For all the time-dependent covariates, a node-specific pre-processing is achieved to summarize the covariate dynamics into a set of time-independent features to be included in the pool of candidates for the splitting (see figure \ref{fig:dynforest}B).
At each node $d$, the trajectory of time-dependent covariate $Y_m \in \mathcal{M}_y^{(d)}$ is modeled using a flexible mixed model \cite{laird_random-effects_1982} as:
\begin{equation}
Y_{ijm} = X^{\top}_{im}(t_{ijm}) \beta_m^{(d)} + Z^{\top}_{im}(t_{ijm}) b_{im}^{(d)} + \epsilon_{ijm}^{(d)} \label{eq:LMM}
\end{equation}

where $Y_{ijm}$ is the covariate value for subject $i$ at time $t_{ijm}$, $X^{\top}_{im}(t_{ijm})$ and $Z^{\top}_{il}(t_{ijm})$ are the $p_m$- and $q_m$-vectors associated with the fixed effects $\beta_m^{(d)}$ and random effects $b_{im}^{(d)}$ (with $b_{im}^{(d)} \sim \mathcal{N}(0,B_m^{(d)})$), respectively. $\epsilon_{ijm}^{(d)}$ denotes the error measurement with $\epsilon_{ijm}^{(d)} \sim \mathcal{N}(0,{\sigma^2_m}^{(d)})$. 

We present the method for continuous Gaussian time-dependent covariates only. However, the pre-processing procedure could be easily adapted to other types of time-dependent covariates by replacing the linear mixed model in \eqref{eq:LMM} by a generalized linear mixed model.

Any specification can be considered for $X^{\top}_{im}(t_{ijm})$ and $Z^{\top}_{im}(t_{ijm})$. To allow for a flexible modeling of the trajectory over time, we consider a basis of natural cubic splines with knots to be determined in input. Although the specification of the model is similar for each node, the maximum likelihood estimation of the parameters is performed at each node on the subset of subjects present at the node (i.e., $\forall i \in \mathcal{N}^{(d)}$). When the covariate had already been selected at a parent node, estimated parameters from the closest parent node are considered as initial values to drastically speed-up the procedure. 

Time-independent features are then derived as the predicted individual deviations to the mean trajectory:

\begin{equation}
\hat{b}_{im}^{(d)} = \hat{B}_m^{(d)} Z^{\top}_{im} \hat{V}^{-1(d)}_{im} (Y_{im} - X_{im} \hat{\beta}_m^{(d)}) \label{eq:RE}
\end{equation}

where $X_{im}$ and $Z_{im}$ are the matrices with $j$-row vectors $X^{\top}_{im}(t_{ijm})$ and $Z^{\top}_{im}(t_{ijm})$ (with $j=1,...n_{im}$),  $\hat{V}_{im}^{(d)} = Z_{im} \hat{B}_m^{(d)} Z^{\top}_{im} + \hat{\sigma}_{\epsilon m}^{(d)} I_{n_i}$, $I_{n_i}$ the $n_i \times n_i$ identity matrix and the hat denotes the Maximum Likelihood Estimates. 

At this stage, the ensemble of candidate features for the time-dependent covariates becomes $\mathcal{M}_{y\star}^{(d)} = \{\hat{b}_{im}^{(d)} ~~ \forall m : Y_m \in \mathcal{M}_y^{(d)} \}$ and the total ensemble of candidate features $\mathcal{M}_\star^{(d)} = \{\mathcal{M}_x^{(d)}, \mathcal{M}_{y\star}^{(d)} \}$ is now only composed of time-independent features.

\subsubsection{Splitting rule} \label{sec:split}

At each node $d \in \mathcal{D}$, the subjects are to be split into the two daughter nodes that are the most different possible according to the outcome (figure \ref{fig:dynforest}D). With survival outcome, the difference is quantified according to the log-rank statistic. In the presence of competing risks, we use instead the Fine \& Gray test statistic \cite{gray_class_1988} which directly quantifies the difference in terms of the probability of the cause of event of interest. 

The splitting procedure requires that each feature $W \in \mathcal{M}_\star^{(d)}$ be dichotomized. For a continuous predictor, this is achieved by considering a dichotomization according to a threshold $c$: $w_i > c$ or $w_i \leq c$. We used each decile of $W$ as a candidate threshold $c$. Alternatively $c$ could be chosen according to values randomly drawn from $W$. For a non-continuous predictor, the dichotomization can be achieved as $w_i \in c$ or $w_i \notin c$ with $c$ each possible subset of $W$ modalities. 

The log-rank test statistic with one cause or Fine \& Gray test statistic with multiple causes is computed for all potential dichotomized features (defined by couple $\{W,c\}$), and the dichotomized feature ($\{W_0^d,c_0^d\}$) that maximizes the test statistic is selected to create the left and right daughter nodes, denoted nodes $2d$ and $2d + 1$, respectively.

\subsubsection{Stopping criteria}

Criteria need to be established to end the recursive procedure of a tree construction. We distinguish two criteria to pursue with the splitting of a node: (i) A minimum number of events called \textit{minsplit}; (ii) A minimum number of subjects in each of the daughter nodes called \textit{nodesize}. These two parameters control the depth of the trees. They need to be carefully determined as a trade-off between the performances of the random forest (with the deeper the trees, the lower the error of prediction) and the computational time. In our examples, we often used $nodesize = 3$ and $minsplit = 5$. When a stopping criterion is reached, the node is considered as a terminal node or leaf $h \in \mathcal{H}$.

\subsubsection{Leaf summary}

The subjects classified in the same leaf are supposed to be homogeneous in terms of their probability of event of interest. Each leaf $h^b$ of tree $b$ is thus summarized by the cumulative incidence function (CIF) for cause $k$ ($k=1,...,K$):

\begin{equation}
\pi_k^{h^b}(t) = P ( T_i < t , \delta_i = k ~|~ i \in h^b) ~,~ \forall t \in \mathbb{R}^+
\end{equation}

An estimate $\hat{\pi}_k^{h^b}(t)$ of the CIF $\pi_k^{h^b}(t)$ is given by the Aalen-Johansen estimator.

\subsection{Individual prediction of the outcome} \label{sec:rf}

\subsubsection{Out-of-bag individual prediction} \label{sec:OOBind}

Let us consider an individual $\star$ with the $P$-vector of time-independent covariates $X_{\star}$ and the ensemble of time-dependent covariate observations $\mathcal{Y}_{\star} = \{ Y_{\star jm}, m=1,...Q, j=1...n_{\star m} \}$. The individual-specific CIF for individual $\star$ in tree $b$ is given by:

\begin{equation}
    \begin{array}{rll}\label{eq:predBtree}
\pi_{\star k}^b (t) &=& P (T_\star < t , \delta_\star = k ~|~ \mathcal{Y}_\star, X_\star, b)\\
&=& P (T_\star < t , \delta_\star = k ~|~ \star \in h_\star^b)\\
&=& \pi_k^{h_\star^b}(t)
\end{array}
\end{equation}

where $h_\star^b$ is the leaf in which individual $\star$ ends when dropping into tree $b$. Specifically, at each node $d$, subject $\star$ is recursively assigned to the left or right node according to whether $w_\star > c_0^d$ or $w_\star \leq c_0^d$. In the case where $W_0^d$ is a predicted random-effect from time-dependent covariate $m$, the random-effect prediction for individual $\star$, $\hat{b}_{\star m}^{(d)}$, is computed using formula \eqref{eq:RE} with the estimated parameters obtained at this specific node $d$.

An ensemble estimate of the individual CIF $\hat{\pi}_{\star k}(t)$ for cause $k$ can finally be defined by aggregating the tree-specific individual predictions $\hat{\pi}_{\star k}^b (t) = \hat{\pi}_k^{h_\star^b}(t)$ over all the trees $\mathcal{O}_\star \subset \{1,...,B\}$ for which $\star$ is $OOB$, as:
\begin{equation} \label{eq:RSFindCIF}
\hat{\pi}_{ \star k}(t) = \frac{1}{|\mathcal{O}_\star|} \sum_{b \in \mathcal{O}_\star} \hat{\pi}_k^{h_\star^b}(t)
\end{equation}

where $|\mathcal{O}_\star|$ denotes the length of $\mathcal{O}_\star$ and $\hat{\pi}_k^{h_\star^b}(t)$ is the Aalen-Johansen estimator in leaf $h_\star^b$ of the $b$-th tree.

\subsubsection{Individual dynamic prediction from a landmark time} \label{sec:dyn_pred}

The methodology described in the previous section for an out-of-bag individual can be used to provide the individual dynamic prediction of the outcome of cause $k$ from the information collected up to a landmark time $s$. Let consider a new subject $\star$ still at risk of the event at time $s$. The covariate information available at the time of prediction is the $P$-vector of time independent covariates $X_\star$ and the history of time-dependent covariates observations up to time $s$ $\mathcal{Y}_{\star}(s) = \{ Y_{\star jm}, m=1,...Q, j=1...n_{\star m}, t_{\star jm} < s \}$. The probability of experiencing cause $k$ of event at a horizon time $w$ is then defined as: 
\begin{equation}\label{eq:inddynpred}
\begin{array}{rll}
\pi_{\star k}(s,w) &=& P(s <T_\star \leq s+w, \delta_\star = k \big| T_\star>s , \mathcal{Y}_\star(s) , X_\star)\\
&=& \dfrac{\pi_{\star k}(s+w)- \pi_{\star k}(s)}{1 - \sum_{l=1}^K \pi_{ \star l}(s)}
\end{array}
\end{equation}
where each $\pi_{\star k}(t)$ (for $k=1,...,K$) can be estimated using equation \eqref{eq:RSFindCIF} with the history of the time-dependent covariates $\mathcal{Y}_{\star}(s)$ up to the landmark time $s$ only, and $\mathcal{O}_\star = \{1,...,B\}$.

\subsection{Error of prediction} \label{sec:OOB_err}

The error of prediction can be used in RSF with two objectives:(i) tuning the hyper-parameters of the RSF (\textit{mtry}, \textit{minsplit} and \textit{nodesize}) to achieve an optimal RSF. This is done by minimizing the OOB error of prediction; (ii) assessing the predictive performances of the optimal RSF. This is achieved by computing the error of prediction for an external validation sample, that is a sample where subjects are OOB for all the trees. 
In this work, we considered mainly the Brier Score measure \cite{blanche_quantifying_2015}, and its integrated version (IBS) between two time points $\tau_1$ and $\tau_2$ to assess the error of prediction.

\subsubsection{IBS for optimizing the RSF} \label{sec:IBS}

For the optimization of the RSF, the IBS estimator is given by $\hat{IBS}(\tau_1;\tau_2) = \int_{\tau_1}^{\tau_2} \hat{BS}(t) \,dt $ with the Brier Score estimated by:
\begin{equation} \label{eq:BS}
\hat{BS}(t) = \frac{1}{N} \sum_{i=1}^{N}  \hat{\omega}_i(t) \Big\{ I(\tilde{T}_i \leq t, \delta_i = k) - 
\hat{\pi}_{ik}(t) \Big) \Big\}^2
\end{equation}

where $\hat{\pi}_{ik}(t)$ is the estimated probability of event of cause $k$ given $Y_i$ and $X_i$ defined in \eqref{eq:RSFindCIF}, and $\hat{\omega}_i(t)$ are Inverse Probability of Censoring Weights (IPCW) that account for the censoring between $\tau_1$ and $\tau_2$ \cite{gerds_consistent_2006}. We used in this work the Kaplan-Meier estimator to compute the probability of censoring in IPCW.

By default, $(\tau_1,\tau_2)$ corresponds to the span of the time to event data. 

\subsubsection{External assessment of RSF predictive performances} 

For the external evaluation of the RSF performances, the IBS computation slightly differs. First, it is now computed on an external sample of size $N^*$, and the information considered is now the information up to the prediction time $s$, with $s \leq \tau_1$, so that $\hat{IBS}^s (\tau_1;\tau_2) = \int_{\tau_1}^{\tau_2} \hat{BS}^s(t) \,dt $ with:
\begin{equation}
 \hat{BS}^s(t) = \frac{1}{N^*} \sum_{\star=1}^{N^*}  \hat{\omega}^s_\star(t) \Big\{ I(\tilde{T}_\star \leq t, \delta_i = k) - 
\hat{\pi}_{\star k}(s,t-s) \Big) \Big\}^2 
\end{equation}

where $\hat{\pi}_{\star k}(s,t-s)$ is the estimated probability of event of cause $k$ between $s$ and $t$ given the information on $Y_\star$ and $X_\star$ up to $s$ (see definition in \eqref{eq:inddynpred}), and $\hat{\omega}^s_\star(t) = \hat{\omega}_\star(t) I(\tilde{T}_\star > s)$.

This external validation step can be incorporated into a k-fold cross-validation strategy. This is what was done in the application in the absence of an actual external dataset, and repeated 50 times to account for the k-fold cross-validation variability. 

\subsection{Importance of the predictors}

Beyond the overall predictive performance of the approach, one can be interested in identifying which predictors are the most predictive. We propose to evaluate the association between event and predictors through two measures: the variable importance and the minimal depth.

\subsubsection{Variable importance}

The variable importance (VIMP) measures the variable prediction ability by computing the increase in OOB error obtained after breaking the link between a given variable and the event. Such a link is broken by permuting the values of variable $p$ at the individual level when $p$ is time-fixed and at the observation level when $p$ is time-dependent. Then, the VIMP statistic for covariate $p$, called $VIMP^{(p)}$, is the difference between the mean over the trees of OOB errors obtained after permuting the values of covariate $p$ ($\hat{IBS}_b(\tau_1,\tau_2)^{(p)}$ for $b={1,...,B}$) and the mean over the trees of the OOB errors ($\hat{IBS}_b(\tau_1,\tau_2)$ for $b={1,...,B}$):
\begin{equation}
VIMP^{(p)}(\tau_1,\tau_2) = \dfrac{1}{B} \sum_{b=1}^B \hat{IBS}_b(\tau_1,\tau_2)^{(p)} - \dfrac{1}{B} \sum_{b=1}^B \hat{IBS}_b(\tau_1,\tau_2)
\end{equation}

where $\hat{IBS}_b(\tau_1,\tau_2)^{(p)}$ and $\hat{IBS}_b(\tau_1,\tau_2)$ are defined similarly as the IBS by computing the Brier Score (in equation \eqref{eq:BS}) only on $b$-tree OOB subjects and using the estimate of $b$-tree individual prediction $\hat{\pi}_k^{h_\star^b}(t)$ defined under equation \eqref{eq:RSFindCIF}.

Large VIMP value indicates a loss of predictive ability when removing covariate $p$ whereas null VIMP value indicates no predictive ability. Due to the permutation procedure, negative VIMP may be obtained. They are interpreted as null VIMP.

\subsubsection{Grouped variable importance} \label{sec:gVIMP}

Because of the potential correlation between variables, the VIMP computed at the variable level may not always indicate the correct variable-specific predictive ability. To assess the predictive ability of correlated variables, Gregorutti \textit{et al.} \cite{gregorutti_grouped_2015} proposed the grouped variable importance (gVIMP) statistic in standard random forest. It consists in simultaneously noising-up all the variables of a given group. We considered the same methodology for our RSF. The overall gVIMP statistic for group $g \in \{1,...,G\}$ is defined as $gVIMP^{(g)} = \dfrac{1}{B}\sum_{b=1}^B \hat{IBS}_b^{(g)} -  \dfrac{1}{B}\sum_{b=1}^B \hat{IBS}_b$ where $\hat{IBS}_g^{(g)}$ denotes the OOB error obtained when noising-up all the variables from group $g$.

\subsubsection{Average minimal depth}

The minimal depth of a variable in a tree corresponds to the distance between the root node and the first node that used the variable for splitting the data. The minimal depth can be averaged across all the trees allowing to rank the predictors. Indeed, during the tree building, the most predictive variables are expected to be chosen for the first splits so the closer the average minimal depth from one, the better the predictive ability of the variable. 
When only a random subset of variables are considered at each node ($mtry < P + Q$), the interpretation of the minimal depth may be blurred. We thus recommend to compute this statistic only for the maximal $mtry = P+Q$. Moreover, because the depth of the trees may vary and some predictors may not be systematically used in the tree building process, we recommend to report the number of trees where the predictor was selected along with the average minimal depth. Note that, compared to the VIMP, the minimal depth can be computed both at the summary feature and at the covariate level allowing to fully understand the tree building process.

\section{Simulation study}

We carried out a simulation study to illustrate the behaviour of \texttt{DynForest} in comparison with alternative methods under two scenarios:
\begin{itemize}
    \item repeated data of two longitudinal predictors. We compared the performances of \texttt{DynForest} with a JM estimated using \texttt{JMBayes} R package \cite{rizopoulos_r_2016}. For \texttt{JMBayes}, we considered the same specification for the linear mixed models as in \texttt{DynForest} and modelled the association with the event with a proportional hazard model (baseline risk function approximated by four cubic splines) that included the current levels and current slopes of the two predictors as covariates. 
    \item repeated data of 20 longitudinal predictors. We could not compare with a JM anymore. Instead, we compared \texttt{DynForest} with a RC technique in which the exact same specification for the linear mixed models and the exact same strategy for the RSF were considered. The difference in the RC was that the linear mixed models were estimated once and for all prior to the application of standard RSF. 
\end{itemize}
For both scenarios, we additionally included two time-fixed predictors unrelated to the event. Finally, we compared the predictive performance of the techniques in predicting the clinical event occurrence at two horizon times $w = 1,2$ from two landmark times $s = 2,4$. We measured the performance with both Brier Score (defined in \eqref{eq:BS}) and Area Under the ROC curve (AUC) with estimators adapted to dynamic prediction \cite{blanche_quantifying_2015}.

\subsection{Design}

For both scenarios, $R = 250$ samples of $N = 500$ individuals were built for the learning step and a single external validation sample of $N = 500$ individuals was generated for evaluating the predictive performance. The generation procedure is detailed in supplementary material (simulation section and tables S1/S2) and summarized below. Individual trajectories are also displayed in supplementary material (figure S1).

For each subject, we generated two time-fixed covariates (one continuous according to a standard Gaussian distribution and one binary according to a Bernoulli with probability 0.5). We also generated repeated data of 2 or 20 continuous time-dependent predictors, for small and large dimension scenarios, respectively. Times of measurement were at baseline and then randomly drawn (using an exponential departure) around theoretical annual visits up to 10 years. Each marker trajectory followed a latent class linear mixed model \cite{proust-lima_joint_2014} with four classes and either class-specific linear individual trajectories or class-specific nonlinear individual trajectories. The risk of event was then generated using a proportional hazard model with a Weibull baseline hazard with shape and scale parameters equal to 0.1 and 2, respectively. For both scenarios, we considered two sub-scenarios according to the form of the dependence function between the predictors and the risk of event using in the linear predictor of the survival model: (i) random-effects and two-by-two interactions between random-effects; (ii) latent class membership directly. 

\subsection{Results}

\subsubsection{Small dimension scenario}

\begin{figure}[H]
  \centering\includegraphics[width=\textwidth]{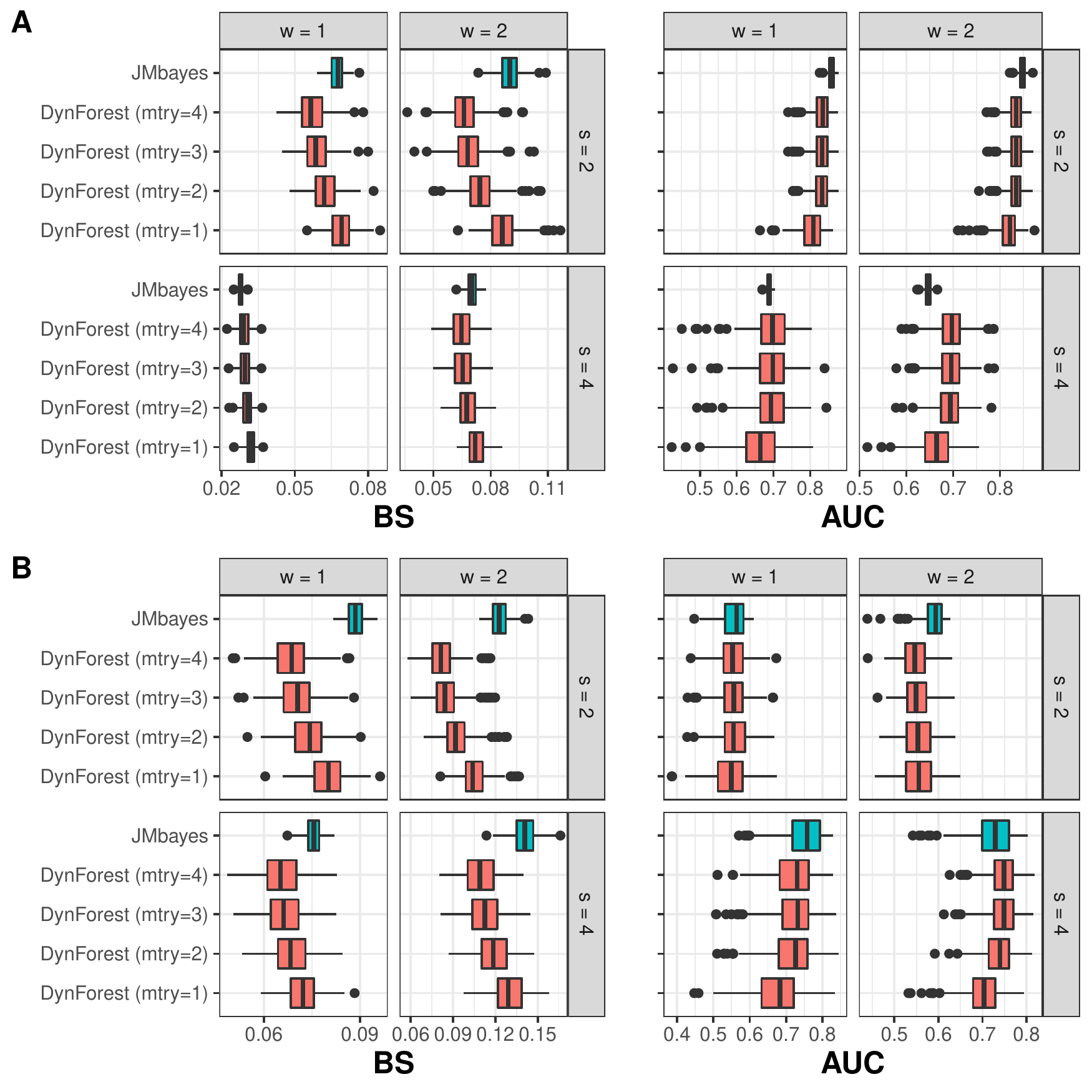}
  \caption{External predictive performances of \texttt{DynForest} and \texttt{JMbayes} in the small dimension scenario of simulations (2 predictors) for the 250 replications. Are reported the Brier Score (BS) and the Area Under the ROC Curve (AUC) at two landmark times $s=2,4$ and two horizons $w=1,2$. The generated joint model included non-linear association between the markers and the event, using random-effects with two-by-two interaction (A) or latent class membership (B). In these scenarios, \texttt{JMbayes} is considered as misspecified. For \texttt{DynForest}, we fixed $nodesize = 3$ and $minsplit = 5$, and their results are reported for all mtry values to underline the importance of this tuning parameter.}
  \label{fig:simuB2}
\end{figure}

Predictive performances on the external dataset are reported in terms of BS and AUC in figure \ref{fig:simuB2}. For \texttt{DynForest}, we fixed $nodesize = 3$ and $minsplit = 5$ whereas we chose to report the results with each possible value of \textit{mtry} to underline the importance of this tuning parameter. As expected, the results varied substantially according to its value. The best performances in terms of BS (minimal BS) was systematically obtained with the largest \textit{mtry}, that is four (two time-dependent and two time-fixed predictors), and the worse with $mtry = 1$. For the AUC, the differences were less visible. In comparison with the JM estimated with \texttt{JMbayes}, \texttt{DynForest} showed overall better predictive abilities, in particular in terms of Brier Score. Again, conclusions were more tempered when considering AUC. It should be noted that JM in this simulation was misspecified in terms of both the individual deviation from the mean trajectory (since simulated using latent classes), and of dependence association since it has been generated as a non-linear function. The objective here was to illustrate that even in small dimension scenarios, \texttt{DynForest} could already be of interest and constituted a competing alternative for individual dynamic prediction purpose.

\subsubsection{Large dimension scenario}

\begin{figure}[H]
  \centering\includegraphics[width=\textwidth]{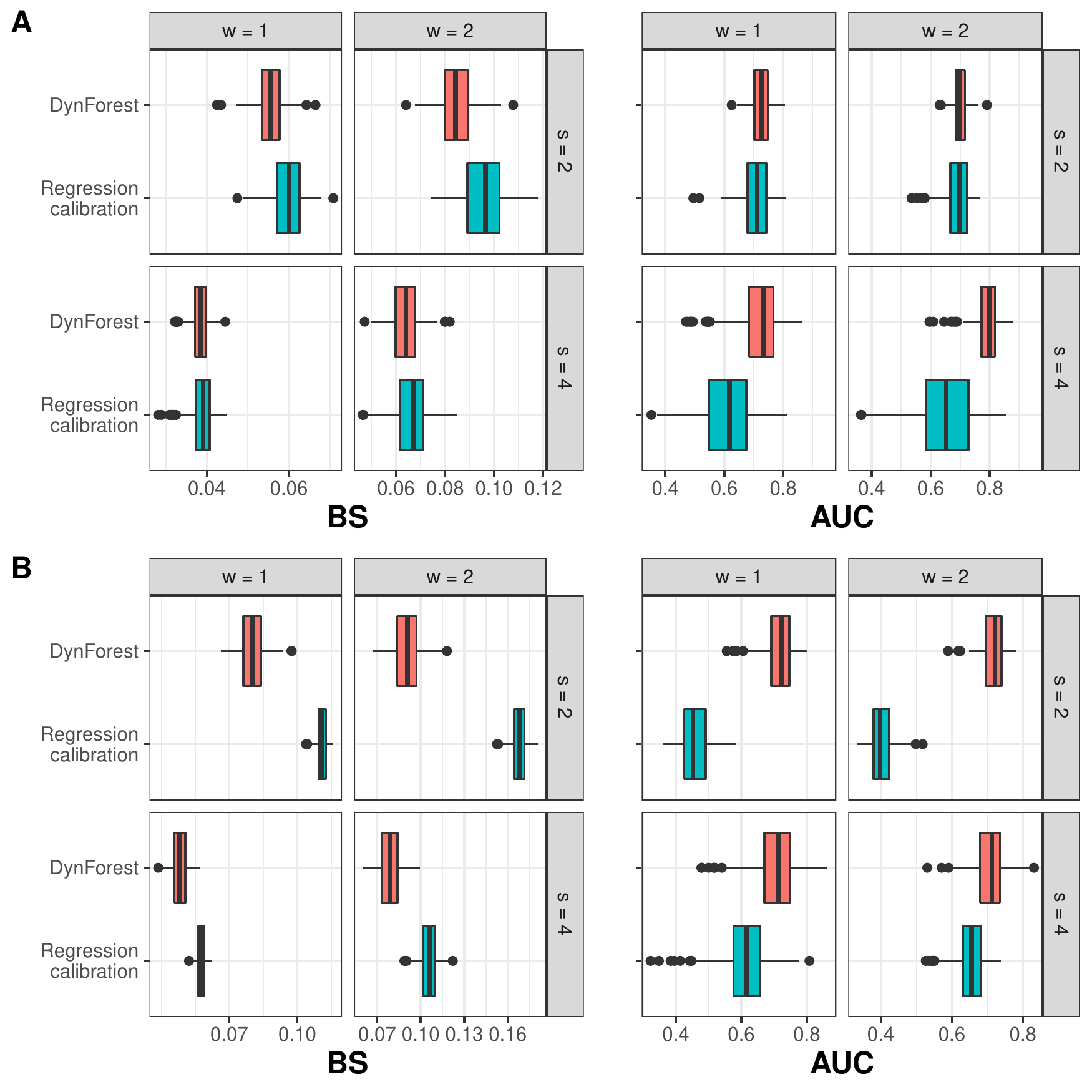}
  \caption{External predictive performances of \texttt{DynForest} and its regression calibration version in the large dimension scenario of simulations (20 predictors) for the 250 replications. Are reported the Brier Score (BS) and the Area Under the ROC Curve (AUC) at two landmark times $s=2,4$ and two horizons $w=1,2$. Non-linear association between the markers and the event was displayed using random-effects with two-by-two interactions (A) or latent class membership (B). The regression calibration version of \texttt{DynForest} consisted in summarizing the time-dependent markers into time-fixed features once for all prior to inclusion in the RSF. We fixed $nodesize = 3$ and $minsplit = 5$. $mtry$ parameter was also fixed for all replications after tuning process on an unique dataset.}
  \label{fig:simuD2}
\end{figure}

In the large dimension scenario, we report in figure \ref{fig:simuD2} the predictive performances on the external dataset of \texttt{DynForest} and its RC counterpart (or two-stage counterpart). For both techniques, we fixed $nodesize = 3$ and $minsplit = 5$ whereas $mtry$ parameter was tuned. Regarding the range of possible values of $mtry$ (from 1 to 22 for \texttt{DynForest} and from 1 to 62 for RC method), tuning this parameter on each dataset was too computationally intensive. Thus, we decided to tune this parameter using an unique dataset, and used the optimal value for all the replications. With non-linear association using random-effects plus interactions, optimal values were $mtry = 9$ and $mtry = 46$ for \texttt{DynForest} and RC, respectively. They were $mtry = 5$ and $mtry = 6$ when linking the markers to the event through the latent class membership. \texttt{DynForest} outperformed the RC technique for both BS and AUC with non-linear association using latent class membership (figure \ref{fig:simuD2}B). Under non-linear association using random-effects with interactions (figure \ref{fig:simuD2}A), the results were still slightly in favor of \texttt{DynForest}. This underlines the substantial impact of not including the time-dependent predictor modeling step within the survival tool construction to correctly account for the correlation between the longitudinal and survival processes and the informative dropout.

\section{Application}

We aimed at predicting the individual probability of dementia in the elderly in the presence of competing death by leveraging the history of repeated data on clinical exam, neuropsychological battery and brain Magnetic Resonance Imaging (MRI) exam. We relied for this on the Three-City (3C) cohort study \cite{3c_study_group_vascular_2003}. 

\subsection{The 3C study}

The 3C study is a French prospective population-based cohort study which enrolled individuals aged 65 years and older from electoral rolls in three French cities (Bordeaux, Dijon and Montpellier). Extensive follow-up interviews were conducted at baseline and then 2, 4, 7, 10, 12, 14 and 17 years after the enrollment including an extensive clinical and neuropsychological exam done in-person at home by a trained psychologist. At 1, 4 and 10 years, a subsample underwent an additional MRI exam. The diagnosis of dementia relied on a two-step
procedure with suspected cased of dementia examined by a clinician and validated by an independent expert committee of neurologists and geriatricians. Deaths were continuously recorded but were considered as a competing event for dementia only in the three years after a negative diagnosis.
Our analytical sample included all the individuals free of dementia at baseline and with at least 1 measure at each of the 29 predictors under study during the follow-up in Bordeaux and Dijon cities. This lead to a sample of $N = 2140$ subjects (with 10766 observations) among which 234 were diagnosed with an incident dementia and 311 died before any dementia (figure S5 in supplementary material).

We considered a total of 24 time-dependent and 5 time-fixed predictors structured into 9 groups: socio-demographic (time-fixed age at baseline, education, gender), cardio-metabolic factors (three time-dependent markers with body mass index, diastolic and systolic blood pressure, and one time-fixed with diabetes status at baseline), medication (time-dependent number of medication), depressive symptomatology (one time-dependent scale of depressive symptomatology), cognition (four time-dependent cognitive tests), functional dependency (one time-dependent scale of instrumental activities of daily living), genetic (time-fixed APOE4 allele carrier status), neurodegeneration (eight time-dependent brain MRI markers including regional volumes and global measures) and vascular brain lesions (six time-dependent markers of white matter hyperintensities). Complete information on the predictors are provided in table S3 to S8 in supplementary material. For longitudinal predictors, individual trajectories are displayed in Figures S2, S3 and S4 in supplementary material.

\subsection{\texttt{DynForest} specification}

The probability of dementia was predicted according to time from the enrollment. MRI data were collected 1.7 times on average and modeled using quadratic and linear trajectories at the population and at the individual level, respectively. Other time-dependent predictors were measured 5.1 times on average. Their trajectories according to time in the study were modeled using natural splines with one internal knot both at the population and individual level. To satisfy the normality assumption of the linear mixed model, all time-dependent predictors were previously normalized using splines transformations \cite{proust-lima_are_2019}.

In the absence of an external dataset available with the same longitudinal predictors and the same target population, predictive abilities were assessed using a 10-fold cross-validation procedure to avoid over-fitting. For each of the 10 folds, \texttt{DynForest} was trained on the sample that excluded the fold (learning step on 90\%) and individual probabilities of dementia were computed on the fold (prediction step on the remaining 10\%). The cross-validation procedure was repeated $R = 50$ times to appreciate the variability of the results.
During the learning step, we systematically fixed parameters $minsplit = 5$ and $nodesize = 3$ to favor deep trees. The $mtry$ parameter was tuned within the range of possible value (from 1 to 29 predictors) to minimize the OOB IBS. On the total sample, we first observed that the OOB IBS decreased rapidly with increasing $mtry$ until a stabilization around $mtry = 15$ (figure S6 in supplementary material). So for each fold, we ran \texttt{DynForest} twice with $mtry = 15$ and $mtry = 20$ and selected the optimal $mtry$ according to the OOB IBS. For the prediction step, individual dementia probabilities were computed for the remaining fold following \eqref{eq:RSFindCIF}.

\subsection{Results}

\begin{figure}[H]
  \centering\includegraphics[width=0.95\textwidth]{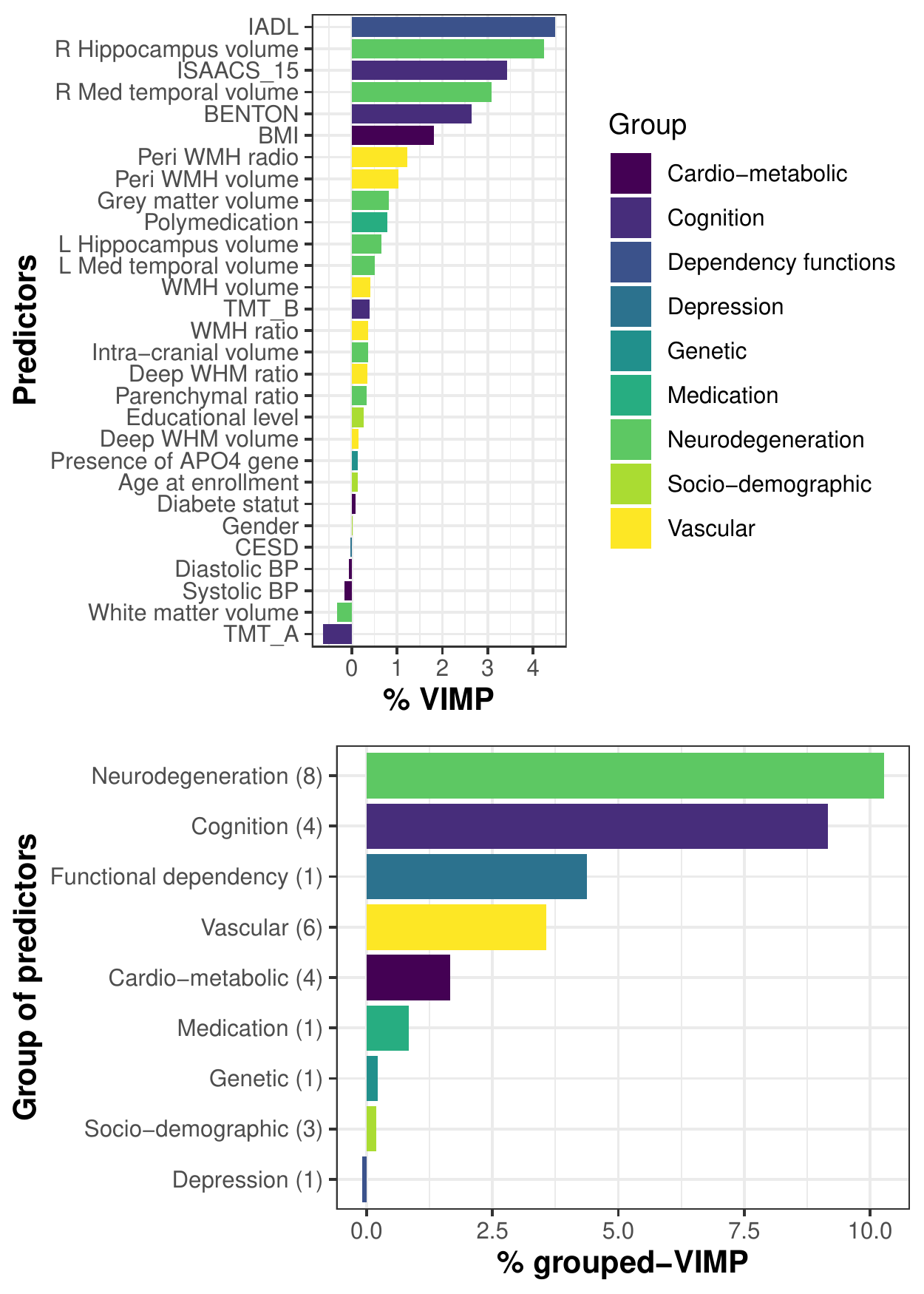}
  \caption{(A) Importance variable (VIMP) and (B) grouped importance variable (gVIMP) averaged over 10 permutation procedures for each dementia predictor or group of dementia predictors. Application in the 3C study.}
  \label{fig:3C_VIMP}
\end{figure}

To better understand the importance of each predictor, we report the VIMP statistics in figure \ref{fig:3C_VIMP}A. The VIMP statistics were computed 10 times and averaged across the replications to reduce the variability due to the permutation procedure. IADL (functional dependency) was the marker the most associated to dementia with a mean gain in IBS of 4.5\%, followed by neuro-degeneration markers with the right hippocampus and lobe medio-temporal volumes (gains of 4.2\% and 3.1\%, respectively), and cognitive tests with the Isaacs Set Test and Benton test (gains of 3.4\% and 2.6\%, respectively). Since the VIMP may not correctly translate the importance of correlated variables, we also reported in figure \ref{fig:3C_VIMP}B the gVIMP grouped by dimensions. The eight neuro-degeneration predictors reached a mean gain of 10.3\% of IBS, and the four cognitive tests a mean gain of 9.2\%. Then, we observed less importance for the unique marker of functional dependency (mean gain of 4.5\%) followed by the six markers of vascular brain lesions (mean gain of 3.6\%).

\begin{figure}[H]
  \centering
  \includegraphics[width=\textwidth]{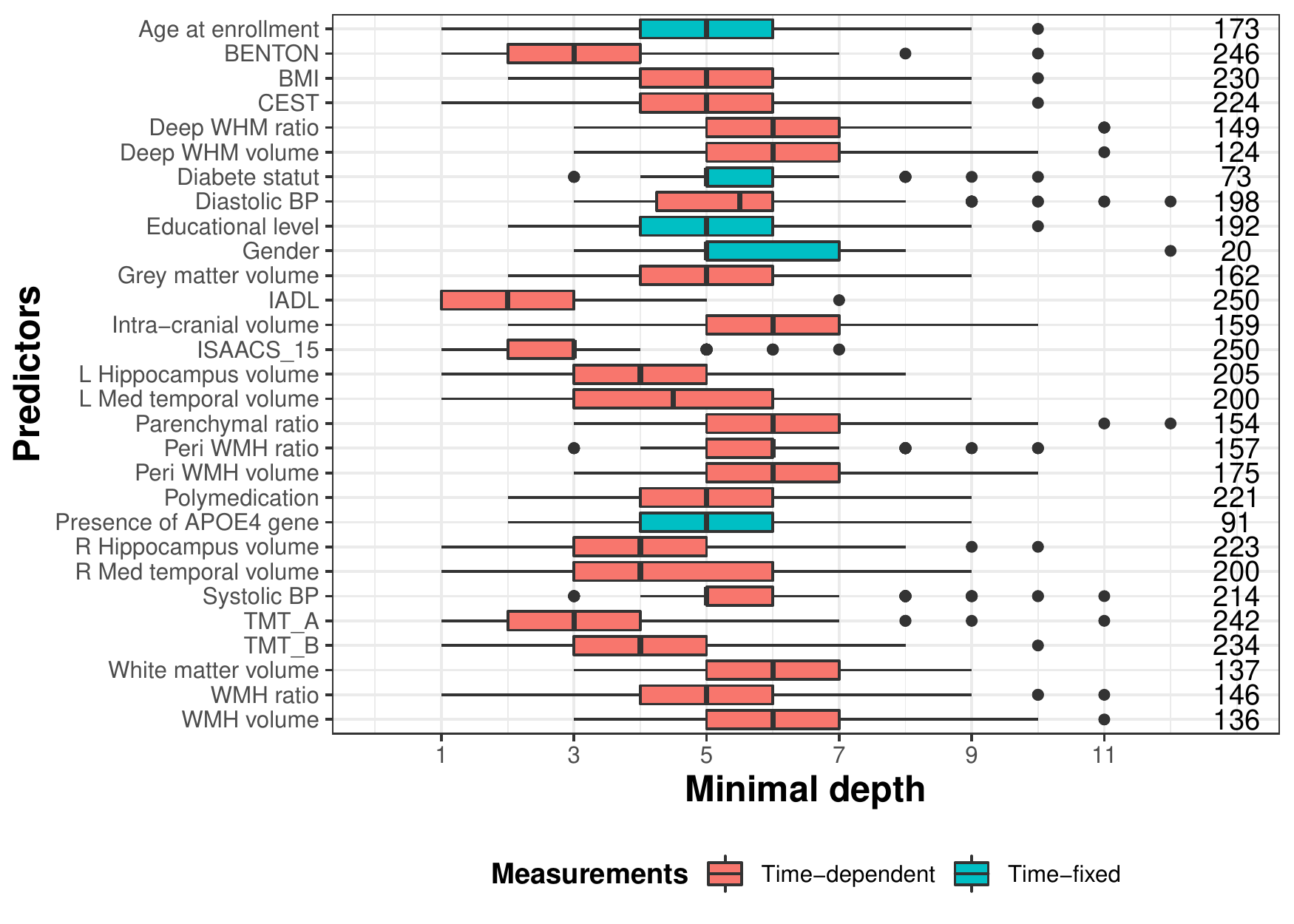}
  \caption{Minimal depth computed with the largest $mtry$ hyper-parameter (i.e. $mtry = 29$) for each predictor of dementia. We display on the right of the graph the amount of tree where the predictor is found among the 250 trees used to build the random forest.}
  \label{fig:3C_mindepth}
\end{figure}

We also computed the minimal depth when using the largest $mtry$ hyper-parameter (i.e. $mtry = 29$) (figure \ref{fig:3C_mindepth}). IADL (functional dependency) and cognition tests (Isaacs Set Test, Benton test and Trail Making Test A) were the predictors with the lowest average minimal depth, and were selected 100\%, 100\%, 98\% and 97\% among the trees, respectively. It means that these predictors were the most effective to split the individuals into homogeneous subgroups according to their risk difference. Except for Trail Making Test A, these results were in accordance with those obtained using the VIMP statistic.

\begin{figure}[H]
  \centering\includegraphics[width=\textwidth]{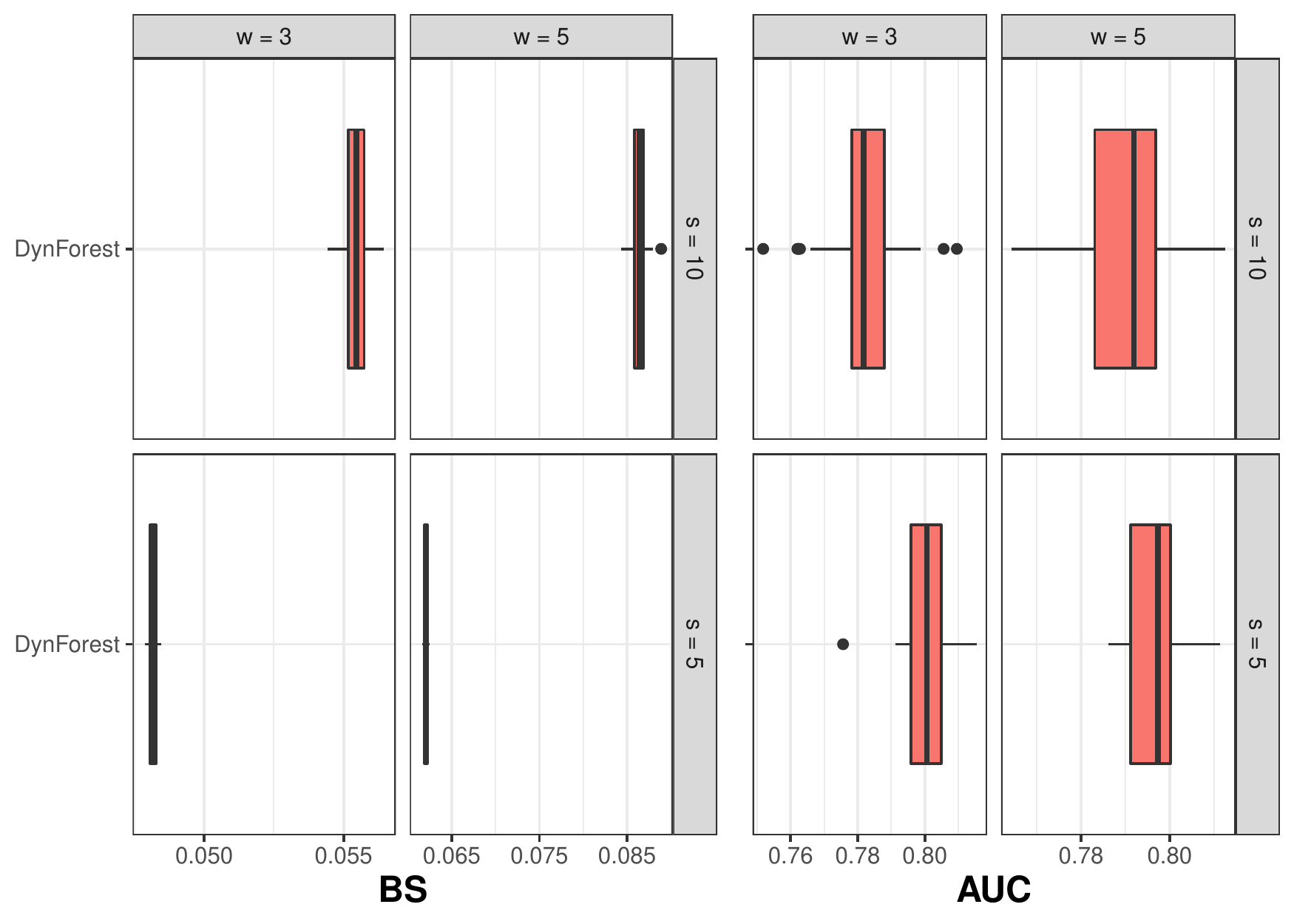}
  \caption{Predictive assessment of dementia at landmark times $s = 5,10$ years and horizon times $w = 3,5$ years using Brier Score (BS) and Area Under the ROC Curve (AUC). Application in the 3C study.}
  \label{fig:3C_BS_AUC}
\end{figure}

We then considered two landmark times $s = 5, 10$ years to assess the predictive abilities of \texttt{DynForest} to predict dementia between $s$ and $s + w$ (horizon times $w = 3,5$ years) from individual history up to time $s$. This resulted in 1727 and 1150 individuals still at risk of dementia, respectively. The cross-validated AUC and BS (figure \ref{fig:3C_BS_AUC}) varied from 0.78 to 0.80 and from 0.048 to 0.086 depending on the landmark and the horizon times.

We finally explored the predictive ability of each predictor in this landmark context by computing the VIMP and gVIMP using only the information prior to 5 years and considering a short span from 5 to 10 years (figure S8). Again, IADL had the largest VIMP value, followed by the Isaacs Set Test and the right hippocampus volume.

\section{Discussion}

We developed an original methodology, called \texttt{DynForest}, to compute individual dynamic predictions from multiple longitudinal predictors. We extended the RSF (which were limited so far to time-fixed predictors) \cite{ishwaran_random_2008, ishwaran_random_2014} to handle endogenous longitudinal predictors. This was achieved by including in the tree building a node-specific internal processing to translate the longitudinal predictors into time-fixed features. \texttt{DynForest} can be used to compute individual dynamic predictions of events as well as quantify the importance of the longitudinal predictors using VIMP and grouped-VIMP adapted to longitudinal data.

Through a simulation study, we first showed in a small dimensional context that \texttt{DynForest} could be a relevant alternative to the JM reference technique. Indeed, in contrast with JM, \texttt{DynForest} does not need to pre-specify the association structure with the event, and may account for nonlinear associations and interactions.  In the second scenario, we considered a larger dimensional context, with 20 longitudinal markers, for which JM could not be estimated anymore. We showed, in this larger dimensional scenario, that \texttt{DynForest} outperformed the RC alternative proposed in the literature \cite{li_dynamic_2019,jiang_functional_2021,lin_functional_2021}. Indeed, in contrast with RC technique, \texttt{DynForest} accounts for the truncation of the repeated data due to the event by re-estimating the mixed models at each node on the node-specific subsample. Since these subsamples become more and more homogeneous regarding the event, the missing at random assumption of the mixed models becomes more and more valid.

Compared with the other methodologies adapted to the large dimensional and longitudinal context, our methodology has the assets of (i) using all available information when landmark approaches \cite{devaux_individual_2022,tanner_dynamic_2021} only include subjects still at risk at landmark time, resulting in a lack of efficiency \cite{ferrer_individual_2019}; (ii) simultaneously analyzing the longitudinal and time-to-event processes when the other methods based on 2-step RC \cite{li_dynamic_2019,jiang_functional_2021,lin_functional_2021} neglect the association leading to a potential bias in the prediction; (iii) allowing for complex and nonlinear association structures between the predictors and the event; (iv) allowing the analysis of potentially high dimensional data (i.e. hundreds/thousands of predictors). Indeed, the longitudinal markers are independently modeled so that the method could be easily applied no matter the number of longitudinal markers. Finally, we introduced two stopping criteria defining the minimum number of events and of subjects required to proceed to a subsequent split. This allows some leaves to have an homogeneous subsample with no events. 

Our methodology has also drawbacks. First, although it may be applied whatever the number of predictors, the computation time may become extremely long in high dimensional settings, in particular with a large number of candidates $mtry$. Indeed, mixed models are to be estimated at each node of each tree even though we managed to fasten the estimation by using the estimates previously obtained as initial values. Second, we only considered continuous longitudinal markers only. However, other natures of repeated markers (e.g. binary, categorical, counts) could be considered using generalized mixed models instead. Third, we relied on linear mixed models for deriving time-fixed features. Functional principal components analysis \cite{yao_functional_2005} could be considered instead. We leave such development for future research. Finally, although we were able to provide the strength of the association between the predictors and the event using the VIMP and gVIMP statistics, these tools do not inform on the sign of the association.


To conclude, using the framework of the random survival forests combined with mixed models for internally processing longitudinal predictors, we tackled the challenge of predicting an event from a potential high number of longitudinal endogenous predictors. \texttt{DynForest} offers an innovative solution accompanied by a user-friendly R package.

\section*{Software}

\texttt{DynForest} R package is available on CRAN and on GitHub at \url{https://github.com/anthonydevaux/DynForest}. Replications scripts for simulation are available on GitHub at \url{https://github.com/anthonydevaux/dynforest_paper_supp}. The application data are available on specific request to the steering committee of the 3C study.

\section*{Supplementary material}
Supplementary material including the simulation description and additional tables and figures for the simulation and application is available online at XXX.

\section*{Acknowledgements}
We thank Dr. Carole Dufouil (Univ. Bordeaux, Inserm) for providing 3C data Dijon center, and Dr. Louis Capitaine for \texttt{FrechForest} R code used in \texttt{DynForest}. 

\section*{Conflict of interest}
The Authors declares that there is no conflict of interest.

\section*{Funding}
This work was funded by the French National Research Agency (ANR-18-CE36-0004 for project DyMES), and the French government in the framework of the PIA3 ("Investment for the future") (project reference 17-EURE-0019) and in the framework of the University of Bordeaux's IdEx "Investments for the Future" program / RRI PHDS.

\bibliographystyle{bmc-mathphys} 
\bibliography{DynForest_paper.bib}

\section*{Figures}

\end{document}